\pdfoutput=1
\documentclass[10pt,twocolumn,letterpaper]{article}

\usepackage{wacv}
\usepackage{times}
\usepackage{epsfig}
\usepackage{graphicx}
\usepackage{amsmath}
\usepackage{amssymb}
\usepackage{booktabs}

\usepackage{tensor}
\usepackage[caption=false]{subfig}
\usepackage{diagbox}
\usepackage{enumitem}
\usepackage{rotating} 
\usepackage{cases}
\usepackage{bm}
\usepackage{color}
\usepackage{colortbl}
\usepackage{xspace}
\usepackage{graphics}
\usepackage{verbatim}
\usepackage{array}

\usepackage{cite}

\newcommand{\M}[1]{\mathtt{#1}}
\newcommand{\V}[1]{\mathbf{#1}}
\newcommand{\diag}{\textrm{diag}}

\newcommand{\cam}[1]{{\mathcal{#1}}}
\newcommand{\pn}{P}
\newcommand{\g}{G}

\def\gb{Gr{\"o}bner basis\xspace}

\def\gbs{Gr{\"o}bner bases\xspace}

\def\pn{{\texttt{P}}}
\def\g{{\texttt{G}}}

\definecolor{mygray}{gray}{.93}
\definecolor{mygray2}{gray}{.97}

\def\Hk{Heikkil{\"{a}}\xspace}  

\def\wacvPaperID{487} 

\wacvalgorithmstrack   

\wacvfinalcopy 

\ifwacvfinal
\usepackage[breaklinks=true,bookmarks=false]{hyperref}
\else
\usepackage[pagebackref=true,breaklinks=true,colorlinks,bookmarks=false]{hyperref}
\fi

\begin{document}

\title{Partially calibrated semi-generalized pose from hybrid point correspondences}
\author{Snehal Bhayani$^\textrm{1}$ \quad
Torsten Sattler$^\textrm{2}$ \quad
Viktor Larsson$^\textrm{3}$ \quad \\
Janne \Hk$^\textrm{1}$ \quad
Zuzana Kukelova$^\textrm{4}$\\
$^\textrm{1}$Center for Machine Vision and Signal Analysis, University of Oulu, Finland\\
$^\textrm{2}$Czech Institute of Informatics, Robotics and Cybernetics, Czech Technical University in Prague\\
$^\textrm{3}$Computer Vision and Geometry Group, Department of Computer Science, ETH Zürich\\
$^\textrm{4}$Visual Recognition Group, Faculty of Electrical Engineering, Czech Technical University in Prague
}


\maketitle
\thispagestyle{empty}

\begin{abstract}
In this paper we study the problem of estimating the semi-generalized pose of a partially calibrated camera, \ie, the pose of a perspective camera with unknown focal length \wrt a generalized camera, from a hybrid set of 2D-2D and 2D-3D point correspondences. We study all possible camera configurations within the generalized camera system. To derive practical solvers to previously unsolved challenging configurations, we test different parameterizations as well as different solving strategies based on state-of-the-art methods for generating efficient polynomial solvers. We evaluate the three most promising solvers, \ie, the $\mathbf{H}51f$ solver with five 2D-2D correspondences and one 2D-3D correspondence viewed by the same camera inside generalized camera, the $\mathbf{H}32f$ solver with three 2D-2D and two 2D-3D correspondences, and the $\mathbf{H}13f$ solver with one 2D-2D and three 2D-3D correspondences, on synthetic and real data. We show that in the presence of noise in the 3D points these solvers provide better estimates than the corresponding absolute pose solvers.
\end{abstract}


\section{Introduction}
\noindent 
Estimating camera geometry, \ie, absolute or relative pose and internal camera calibration, is a fundamental problem in computer vision with many applications, \eg, in camera calibration~\cite{Schops_2020_CVPR,zhang2000TPAMI}, structure-from-motion (SfM)~\cite{wu2013towards, snavely2006photo, sweeney2015optimizing, schonberger2016structure}, localization~\cite{Sattler2017PAMI,Brachmann2019ICCVa,Sarlin2019CVPR}, visual odometry~\cite{mur2015orb,mur2017orb},  and image retrieval~\cite{Philbin07CVPR,Weyand11ICCV}. 

Camera geometry solvers are usually used inside RANSAC-style hypothesis-and-test frameworks~\cite{fischler1981random}. For efficiency, it is therefore important to employ minimal solvers that generate the solution from a minimal number of point correspondences. Minimal relative and absolute pose problems 
have been extensively studied for decades with many solutions for calibrated cameras~\cite{nister2004efficient,fischler1981random,Persson_2018_ECCV}, partially calibrated cameras with unknown focal length \cite{Stewenius2005,kukelova2017clever,Bujnak08CVPR,Larsson2017ICCV,kukelova2016efficient}, cameras with unknown radial distortion \cite{Larsson2017ICCV, kukelova2015radial, Kukelova13ICCV, ByrodKJPA2008},  
and solutions assuming known gravity direction 
~\cite{saurer2017homography, ding2019efficient,kukelova2010closed}. 
The minimal solvers to these problems are based on different parameterizations and different solution methods. 
\Eg, the absolute pose problem for a camera with unknown focal length has been solved 
based on ratio of distances~\cite{Bujnak08CVPR}, 
the 3.5pt formulations from~\cite{Wu2015, Larsson2017ICCV}, 
and a solution based on the Cayley parameterization of rotation that is solved using the extremely efficient 3Q3 solver~\cite{kukelova2016efficient}.

\begin{figure}[t]
    \centering
	    \includegraphics[width=0.3\columnwidth]{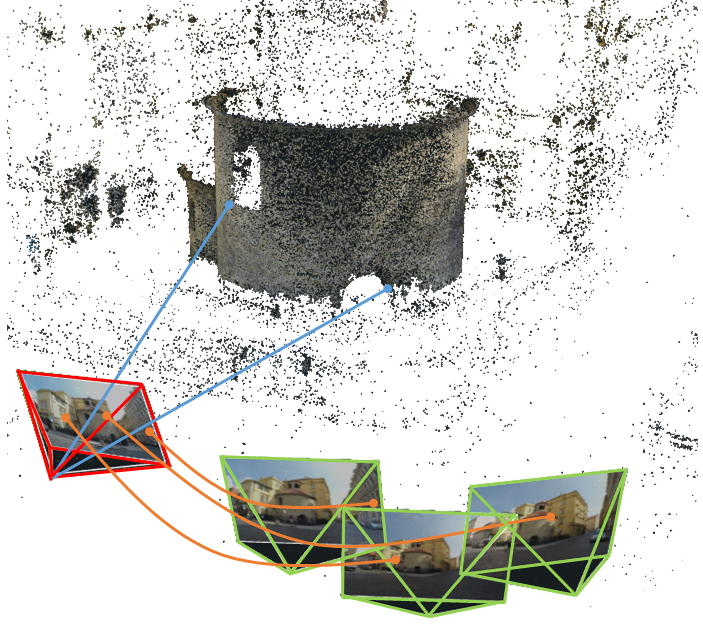}
	    \includegraphics[width=0.6\columnwidth]{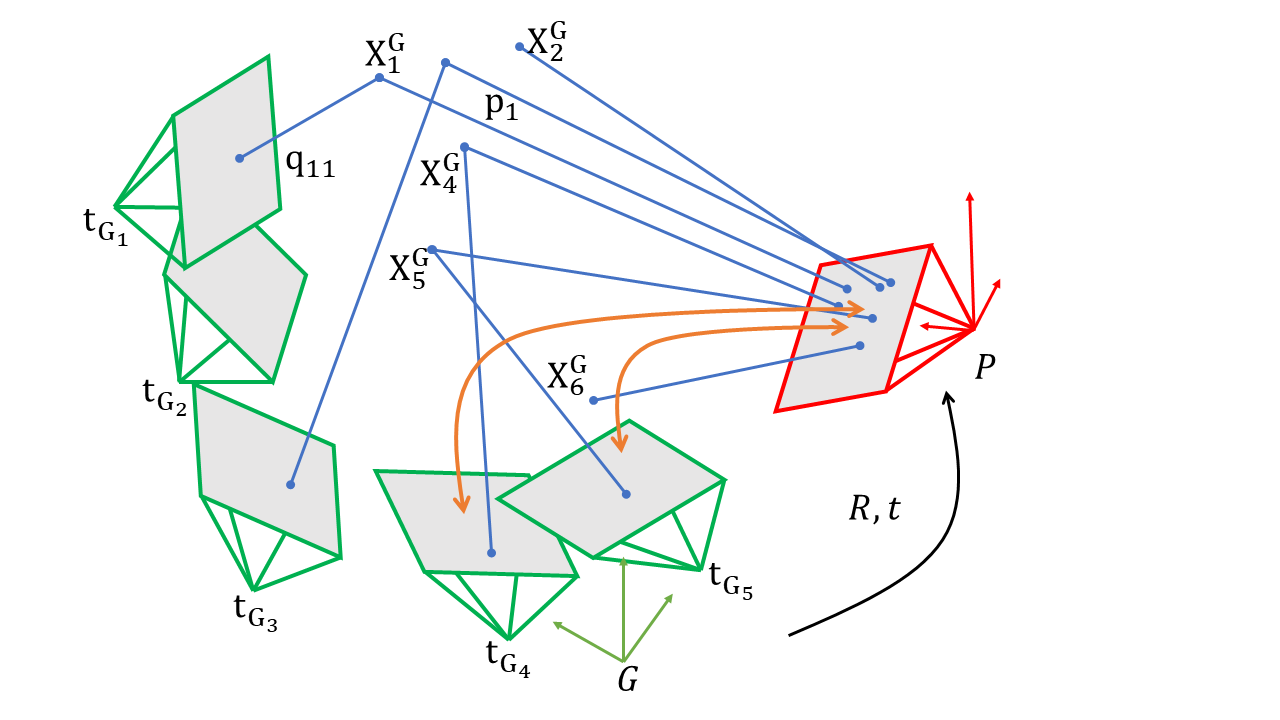}
        \caption{An illustration of the problem configuration. }
    \label{fig:illustration}
\end{figure}

\noindent \textbf{Generalized camera:}
All 
above-mentioned algorithms assume 
the central perspective projection model (potentially with radial distortion). 
Recently, several minimal solutions for generalized cameras were proposed.
A generalized camera~\cite{pless2004camera} can be represented by a arbitrary set of rays with, in general, different projection centers. This camera model has many applications, \eg, in SfM~\cite{Wu_iccv15} and  visual localization~\cite{Wald2020ECCV,Stenborg20203DV}, where we either work with multi-camera systems or  we have to register a new camera (a query image) to a set of cameras with known poses modeled as a generalized camera composed of the known perspective ones. The latter case is called the semi-generalized pose estimation problem. 
In many situations, estimating the camera pose \wrt the generalized camera leads to more accurate poses compared to estimates from pair-wise epipolar geometries, especially thanks to a larger field-of-view of the generalized camera~\cite{sweeney2015computing}. Moreover, in contrast to pinhole epipolar geometry,
 the scale of the translation can be recovered.

While the problem of estimating the absolute pose of a generalized camera can be solved very efficiently using the 3Q3 solver~\cite{kukelova2016efficient} (the final solver runs in a few $\mu$s), the problem of estimating the relative pose of two generalized cameras is significantly more complex~\cite{Stewnius2005SolutionsTM}. This problem results in a complex system of 15 polynomial equations with $64$ solutions and with a \gb solver~\cite{Stewnius2005SolutionsTM} that is infeasible for real-time applications.
Recently, several solutions to different semi-generalized relative pose problems were proposed. 
In~\cite{Wu_iccv15}, the authors considered a semi-generalized epipolar geometry problem, \ie, the problem of estimating the relative pose together with the scale of the translation between one perspective and one generalized camera from 2D-2D correspondences. Due to the complex geometry of the problem, minimal solutions to only four different configurations were presented, \ie, the $\mathbf{\M E_{5+1}}$ and $\mathbf{\M E_{4+2}}$ solvers for calibrated pinhole cameras, and the $\mathbf{\M Ef_{6+1}}$ and $\mathbf{\M Ef_{5+2}}$ solvers for pinhole cameras with unknown focal length. 
Here, a generalized camera consists of multiple perspective ones. 
4+2 denotes a configuration where four point correspondences come from one of these perspective cameras 
and the remaining two correspondences come from 
one or two other perspective cameras. 
The proposed $\mathbf{\M E_{4+2}}$ and $\mathbf{\M Ef_{5+2}}$ solvers were impractical for real-time applications since they perform eliminations of huge matrices and have running times of $1.2 ms$ and $13.6 ms$, respectively.
Recently, \cite{BhayaniSDBHK2021} showed that for planar scenes, the semi-generalized relative pose problem can be, after using suitable parameterizations and after eliminating some unknowns, solved efficiently by finding the roots of a single variable polynomial. 
The authors proposed such efficient minimal solutions for calibrated and partially-calibrated semi-generalized homography estimation and all different configurations of 2D-2D correspondences in a generalized camera.

\vspace{-0.7ex}
\noindent \textbf{Hybrid correspondences:}
In~\cite{JosephsonBKA2007}, the authors suggested to use combinations of 2D-2D and 2D-3D correspondences, \ie, hybrid correspondences, for  visual localization.
Using hybrid correspondences has several advantages. While a 2D-2D correspondence provides only one constraint on the camera geometry, a 2D-3D correspondence provides two constraints and therefore decreases the number of correspondences needed for pose estimation. On the other hand, 3D points may not be available for many 2D detections in the query image, \ie, it is not possible to triangulate these points, or these 3D points can be noisy. Thus, a combination of 2D-2D and 2D-3D correspondences may bring a benefit from both and, as shown in~\cite{camposeco2018hybrid}, can result in better pose estimates.
\cite{JosephsonBKA2007} listed all possible minimal configurations of hybrid point correspondences for semi-generalized pose estimation for calibrated, partially calibrated (unknown focal length) and uncalibrated perspective cameras \wrt a generalized camera. While the authors estimated the number of solutions for all cases, they proposed solvers for only several simple configurations, including configurations where all 2D points are observed by one camera inside the generalized camera or the calibrated case with two 2D-2D and two 2D-3D correspondences. They also suggested a solution to the case of uncalibrated cameras with one 2D-2D correspondences and five 2D-3D correspondences. 

Recently, 
\cite{camposeco2018hybrid} proposed minimal solutions to several problems of estimating the pose of calibrated cameras, and cameras with known vertical direction, from hybrid point correspondences. 
These solutions assume that both cameras are generalized, \ie, a fully generalized case.
The final solvers were mostly obtained using the automatic generator based on \gbs~\cite{kukelova2008automatic}. Together with the proposed minimal solvers for calibrated cameras, a RANSAC-based approach that is automatically selecting the “best” type of solver for each RANSAC iteration was presented. The solver to be used in
the next iteration is selected in a data-driven way using a probability-guided sampling strategy, allowing it to adapt to the quality of the provided correspondences. The paper showed that properly combining different types of correspondences and different camera pose solvers for such correspondences can bring a significant benefit in the performance of RANSAC.
Even though some of the proposed solvers are efficient, even for calibrated cameras there already are configurations that result in large 
solvers, \eg, the solver that uses four 2D-2D and one 2D-3D correspondence and which has to perform elimination of a matrix of size $244 \times 277$.

In this paper, we study challenging unsolved problems for estimating the \textit{semi-generalized pose of a partially calibrated camera}, \ie, the pose of a perspective camera with unknown focal length \wrt a generalized camera, \textit{from a hybrid set of 2D-2D and 2D-3D point correspondences}. The proposed solvers fill the gaps that still remain in the arsenal of minimal solvers and provide new alternatives for pose estimation of a camera with unknown focal length\footnote{Note that in many applications, the only intrinsic parameter of a fully uncalibrated camera that needs to be estimated is the unknown focal length.} that can be efficiently used inside the hybrid RANSAC framework~\cite{camposeco2018hybrid}. We assume a semi-generalized case, compared to the fully generalized case considered in~\cite{camposeco2018hybrid}, since this scenario appears more often in applications, \eg, in visual localization, and results in simpler and faster solvers. The main \textbf{contributions} of the paper are:

\vspace{-1.4ex}
\begin{enumerate}
    \item We propose solutions to all possible minimal point configurations for semi-generalized pose estimation of a partially calibrated camera from a hybrid set of 2D-2D and 2D-3D point correspondences. The proposed solvers include (i) the $\mathbf{H}51f$ solvers with five 2D-2D correspondences and  one 2D-3D  correspondence, (ii) the $\mathbf{H}32f$ solvers with three 2D-2D and two 2D-3D correspondences, and (iii) the $\mathbf{H}13f$ solvers with one 2D-2D correspondence and three 2D-3D correspondences. In all three cases we consider all possible camera configurations within the generalized camera.
    \vspace{-1.0ex}
    \item To derive efficient and stable solvers for all challenging configurations, we test different parameterizations of the problem \eg,  using the essential matrix, homographies, quaternions, and the Cayley parameterization of rotation, as well as different solution strategies based on state-of-the-art methods for generating efficient polynomial solvers, \eg, the elimination ideal method~\cite{kukelova2017clever}, heuristic-based basis sampling approach~\cite{larsson2018beyond} and resultants~\cite{BhayaniKH2021}\footnote{Finding a feasible formulation, \ie, one leading to a practical solution,  and deriving an efficient and stable solver to the resulting polynomial system usually requires many non-trivial ``tricks'' and a good knowledge of both camera and algebraic geometry. Thus, studying different formulations and testing different solution strategies  itself is an important contribution.}.
    \vspace{-1.0ex}
    \item We test the most practical solvers on synthetic as well as real data. We show that in the presence of noise in 3D points and for special type of motions, \eg, forward motion, these solvers provide better estimates than corresponding absolute pose solvers. 
\end{enumerate}

\section{Problem formulation}\label{sec:problem_formulation}
\noindent Let us consider a camera setup as depicted in Fig. \ref{fig:illustration}.  We denote the pinhole query camera as $\cam{P}$ and the generalized camera as $\cam{G}$. 
The generalized camera $\cam{G}$ is assumed to be fully calibrated, and it consists of a set of pinhole cameras denoted as $\lbrace \cam{G}_1, \cam{G}_2, \dots, \cam{G}_k \rbrace$. In this paper, we consider $\cam{P}$ to be partially calibrated. Its calibration matrix is of the form $\M{K} = \diag(f,f,1)$ with unknown focal length $f$. 
We use the upper index to denote a coordinate system. We consider two different coordinate systems for the generalized camera $\cam{G}$: 
the local coordinate system of the generalized camera $\cam{G}$ as a single entity, and the local coordinate
systems of each of the pinhole cameras, $\cam{G}_i$. Let $\M{R}_{\g_i}$, $\V{t}_{\g_i}$ denote the rotation and
translation required to align the local coordinate system of $\cam{G}_i$ to the local coordinate system of the
generalized camera $\cam{G}$. Let $\M{R}_{\g}$, $\V{t}_{\g}$ denote the rotation and translation required to align the
local coordinate system of $\cam{G}$ to the local coordinate system of $\cam{P}$: 
Let $\V X^{\pn} \in \mathbb{R}^3$ and  $\V X^{\g} \in \mathbb{R}^3$ be the coordinates of the point $\V X$ in the local coordinate system of $\cam{P}$ and the local coordinate system of $\cam{G}$, respectively. 
It holds that $\V X^{\pn} = \M R_{\g}\V X^{\g} + \V{t}_{\g}$.

For such a camera setup, our goal is to estimate the rotation $\M{R}_{\g} \in \bf{SO}(3)$ and the translation $\V{t}_{\g} \in \mathbb{R}^3$ between the generalized camera $\cam{G}$ and the perspective camera $\cam{P}$, \ie, to align the local coordinate system of $\cam{G}$ with the local coordinate system of $\cam{P}$. Additionally we also need to estimate the focal length $f$ of the camera $\cam{P}$.  For the sake of brevity, we replace $\M{R}_{\g}$ with $\M{R} $ and $\V{t}_{\g}$ with $\V{t}$. 

Let us assume a 3D point $\V{X}_j$ observed by the perspective camera $\cam{P}$ and the camera $\cam{G}_i$, \ie, the  $i$-th constituent perspective camera from the generalized camera $\cam{G}$. Let us denote the image points detected in $\cam{P}$ and $\cam{G}_i$ as $\V p_j = [x_j, y_j, 1]^\top$ and $\V g_{ij} = [x^{\cam{\g}_i}_j, y^{\cam{\g}_{i}}_j, 1]^\top$, respectively. With this notation, the coordinates of the 3D point $\V{X}_j$ in the local coordinate system of $\cam{P}$ are 
\begin{eqnarray} \label{eq:Xp}
\V{X}_j^{\cam{\pn}} = \alpha_{j} \M{K}^{-1} \V p_j \enspace,
\end{eqnarray}
where $\M{K}$ is the calibration matrix of the camera $\cam{P}$ and $\alpha_j$ represents the depth of the point $\V X_j$ in $\cam{P}$. A similar relationship holds for the coordinates of the 3D point $\V X_j$ in the local coordinate system of $\cam{G}_i$ as 
\begin{eqnarray}
\V{X}_j^{\g_i} = \beta_{ij} \M{K}_{\g_i}^{-1} \V g_{ij} \enspace, \label{eq:XGi}
\end{eqnarray}
where $\M{K}_{\g_i}$ is the calibration matrix of the camera $\cam{G}_i$ and $\beta_{ij}$ represents the depth of the point $\V X_j$ in $\cam{G}_i$. To obtain the relationship between $\V{X}_j^{\pn}$ and $\V{X}_j^{\g_i}$ we have to transform them into the same coordinate system, \ie, in this case the local coordinate system of $\cam{P}$. This gives us 
\begin{eqnarray}
  \M R (\beta_{ij} \M{R}_{\g_i}  \M{K}_{\g_i}^{-1} \V g_{ij} + \V t_{\g_i}) + \V t = \alpha_{j}  \M{K}^{-1} \V p_j \enspace.
\end{eqnarray}
Note that here we use the fact that $\M R = \M{R}_{\g} $  and $\V t = \V t_{\g}$.
Since in our case $\M{R}_{\g_i}, \V{t}_{\g_i}$ and  $\M{K}_{\g_i}$ are known, for better readability we substitute  $\V q_{ij} = \M{R}_{\g_i}  \M{K}_{\g_i}^{-1} \V g_{ij}$ and obtain 
\begin{eqnarray}\label{eq:2d2d}
  \M R (\beta_{ij} \V q_{ij} + \V t_{\g_i}) +\V t =  \alpha_{j} \M{K}^{-1} \V p_j \enspace.
\end{eqnarray}
This denotes the constraint imposed by a 2D-2D correspondence $\V p_j \leftrightarrow (\V q_{ij}, \V t_{\g_i})$. Similarly, if we have a 2D-3D correspondence between a  2D point $\V{p}_j$ and a 3D point $\V X_j^{\g}$ in the local coordinate system of the generalized camera $\cam{\g}$,
then the resulting constraint is 
\begin{eqnarray}\label{eq:3d2d}
\M R \V X_j^{\g} +\V t =  \alpha_{j} \M{K}^{-1} \V p_j \enspace. 
\end{eqnarray}

\section{Minimal solvers }
\noindent The problem of semi-generalized pose from hybrid point correspondences has seven degrees of freedom (d.o.f.), three each for $\M R$ and $\V t$, and one for $f$.
From the constraint~\eqref{eq:2d2d} induced by each 2D-2D point correspondence, we can eliminate the depths $\alpha_j$ and $\beta_{ij}$ to obtain
\begin{eqnarray}\label{eq:2d2d_elim_depths}
( \V p_j )^\top  \left[ \M{K} \M R  \V q_{ij} \right]_{\times} (\M{K} \M R \V t_{\g_i} + \M{K} \V t) =  0 \enspace,
\end{eqnarray} where the notation $\left[ \V{a} \right]_{\times}$ indicates the skew-symmetric matrix of the vector $\V{a} \in \mathbb{R}^3$. Thus, each 2D-2D point correspondence gives us one equation. 
Similarly, from the constraint~\eqref{eq:3d2d} induced by each 2D-3D point correspondence, we can eliminate the depth $\alpha_j$ to obtain 
\begin{eqnarray}\label{eq:3d2d_elim_depths}
\left[  \V p_j \right]_{\times} (\M{K} \M R \V X_j^{\g} + \M{K} \V t) = \V{0} \enspace,
\end{eqnarray} which gives us two linearly independent equations.

A hybrid point configuration $(m,n)$ consists of $m$ 2D-2D 
and $n$ 2D-3D point correspondences. 
It results in a total of $m+2n$ linearly independent equations.
Since in this case we have $7$ d.o.f.,
for any hybrid point configuration $(m,n)$ to lead to a minimal problem, we require that $m+2n=7$. We denote the solver for such a hybrid point configuration $(m,n)$ as $\mathbf{H}mnf$ in this paper.

A given hybrid point configuration $(m,n)$ can have different configurations of the generalized camera $\cam{G}$, based on the largest number $k \leq m$ of 2D-2D correspondences detected  by the same pinhole camera $\cam{G}_i$ within the generalized camera $\cam{G}$. For brevity we denote such a case as $[k]$. 

\noindent \textbf{Parameterizations:}
For all configurations we generated solvers using three different types of parameterizations: 
\vspace{-1.4ex}
\begin{itemize}
    \item \textit{Rotation $\&$ translation} ($\M R \&  \V t$): This parameterization correspond 
    to Eqs.~\eqref{eq:2d2d_elim_depths} and~\eqref{eq:3d2d_elim_depths}. Here we tested Cayley and quaternion-based parameterizations of the rotation matrix $\M R$. For the quaternion representation, inspired by \cite{Zheng2015ICCV}, we use a four variable reparameterization of the product $\M K \M R$. This is important for removing symmetries and halving the number of solutions. We denote the initial polynomial system as $E$. 
\vspace{-1.4ex}
    \item \textit{Homography} ($\M H$): We tested this parameterization for all configurations 
    with at least three 2D-2D correspondences coming from the same camera $\cam{G}_i$\footnote{Note, that here we can project 3D points to the camera $\cam{G}_i$ to obtain a 2D-2D point correspondence. This means that the only case where this parameterization is not applicable is $\mathbf{H}51f[1]$.} or three 2D-3D correspondences. Such three correspondences define a plane and thus a $3 \times 3$ homography matrix $\M{H}$ induced by this plane. Three correspondences give us six linear equations in the elements of $\M{H}$ and thus can be used  to parameterize $\M{H}$ using a three dimensional null space. Here, the initial system $E$ is defined by this parameterization, together with equations coming from the decomposition of the homography matrix $\M{H}$ and the equations from the remaining correspondences.
\vspace{-1.4ex}
    \item \textit{Fundamental matrix} ($\M F$): We tested this parameterization for configurations where we have five or six 2D-2D correspondences coming from the same camera $\cam{G}_i$\footnote{We can again project 3D points to the camera $\cam{G}_i$ to obtain a 2D-2D point correspondence.}. Since each such correspondence gives us a linear constraint on the fundamental matrix $\M F$, we can parameterize it using a four resp. three dimensional null space. Here, the initial system $E$ is defined by this parameterization, together with the equations coming from the decomposition of $\M F$ and those induced by the remaining point correspondences.
\end{itemize}

\noindent To simplify the initial system of equations $E$ in each configuration and parameterization, we tried to eliminate different variables using the elimination ideal method~\cite{kukelova2017clever}. 
We also tried out different transformations of the coordinate systems for the generalized camera $\cam{G}$ and the pinhole camera $\cam{P}$. 
To solve the resulting system of polynomial equations, we used two state-of-the-art algebraic methods for generating efficient polynomial solvers, \ie, the \gb method\cite{larsson2017efficient} including the basis sampling strategy~\cite{larsson2018beyond} and the hidden variable resultant-based method \cite{BhayaniKH2021}. 

This results in huge number of different combinations of different parameterizations, point configurations, solution strategies, and solvers that we generated. In the paper, we present the fastest solvers among all generated solvers for each hybrid point configuration $(m,n)$ and all possible generalized camera configurations $[k]$. These solvers are summarized in Tab.~\ref{tbl:solvers}. 
Note that we have not studied the $(7,0)$ case because our goal in this paper is to study hybrid point configurations with at least one 2D-3D point correspondence.
The first two columns in this table report the solver/problem name and the number of solutions\footnote{Note that for some problems,  the reported number of solutions does not correspond to the number of solutions presented in~\cite{JosephsonBKA2007}. The reason for this is two-fold, first of all~\cite{JosephsonBKA2007} did not consider possible symmetries, and second, in some problems the authors computed solutions to the case where both the perspective and the generalized camera have a common unknown focal length. However, this scenario is impractical.} of the given configuration and its particular formulation, the third column lists the feasible configurations of cameras in a generalized camera that can be solved using the particular solver, the fourth and the fifth columns list the size of the smallest/fastest solver, generated using the basis sampling strategy~\cite{larsson2018beyond} (elimination template matrix size), resp.~the resultant-based method~\cite{BhayaniKH2021} (Generalized Eigenvalue Problem size),
the sixth column is the parameterization of the problem that leads to 
this
solver, and  the last two columns depict the hybrid point configuration.

Next, we describe the fastest solvers for each studied hybrid point configuration, \ie,  $\mathbf{H}13f$, $\mathbf{H}32f$, and  $\mathbf{H}51f$.

\begin{table}[t]
    \centering
    \resizebox{\columnwidth}{!}{
    \begin{tabular}{|c|c|c|c|c|c|c|c|}
    \toprule
        Problem & $\#$sols & Gen. cam. & GB \cite{larsson2018beyond}  & Res. \cite{BhayaniKH2021} & Param. & 2D-2D & 2D-3D \\
        \midrule
        $\mathbf{H}13f$& $12$  & $[1]$  & $70 \times 82$  & $91 \times 91$ & $\M{H}$ & $1$ & $3$ \\
        $\mathbf{H}32f$& $26$  & $[1]$,$[2]$,$[3]$  & $289 \times 315$ & $-$ &  $\M{R} \& \V{t}$ & $3$ & $2$ \\ 
        $\mathbf{H}51f$& $56$  & $[1]$ & $506 \times 562$ & $537 \times 537$  & $\M{R} \& \V{t}$ & $5$ & $1$ \\
        $\mathbf{H}51f$& $50$  & $[2]$,$[3]$ & $511 \times 561$ & $374 \times 374$ & $\M{R} \& \V{t}$ & $5$ & $1$ \\
        $\mathbf{H}51f$& $38$  & $[4]$ & $390 \times 428$ & $243 \times 243$ & $\M{R} \& \V{t}$ & $5$ & $1$ \\
        $\mathbf{H}51f$& $9$  & $[5]$& $9 \times 18$ & $10 \times 10$ & F & $5$ & $1$ \\        
        \bottomrule
    \end{tabular}}
    \vspace{2pt}
    \caption{Partially calibrated semi-generalized pose solvers for hybrid point correspondences. $\mathbf{H}mnf$ denotes a case with $m$ 2D-2D point correspondences and $n$ 2D-3D point correspondences. $[k]$ denotes a case where no more than $k$ 2D-2D correspondences are detected by a camera $\cam{G}_i$ within the generalized camera $\cam{G}$.} 
    \label{tbl:solvers}
\end{table}

\subsection{$\mathbf{H}13f$}
\noindent In this case we have one 2D-2D point correspondence $ \V p_1  \leftrightarrow (\V q_{11}, \V t_{\g_1})$ and  three 2D-3D point correspondences $  \V p_j \leftrightarrow \V X^{\g}_{j}, \ j=2,\dots,4$, and therefore only one hybrid point configuration, \ie, the configuration $[1]$.
For this configuration, the parameterization that led to the smallest solver is the  \textbf{\textit{Homography parameterization}}.


In this case, the three 3D points $\V X^{\g}_{j}, \ j=2,\dots, 4$,  define a plane in the local coordinate system of the generalized camera $\cam{G}$. Let us denote this plane as $\pi$ and its vector as $\V{N}$, encoding both the direction of the plane normal and the distance from the origin. Therefore
\begin{equation}
    \V X^{\g}_j \in \pi \implies \V{N}^\top \V X^{\g}_j + 1 = 0 \enspace , \enspace  j=2,\dots,4 \enspace.
\end{equation}

\vspace{2px}
\noindent \textbf{Coordinate system transform:}
\noindent W.l.o.g., we can rotate and translate the coordinate system of $\cam{G}$ such that $\V{N} = \begin{bmatrix} 0 & 0 & d \end{bmatrix}^\top, \ d \neq 0$ and $\V{t}_{\g_1} = \begin{bmatrix} 0 & 0 & 0 \end{bmatrix}^\top$.
W.l.o.g., we can also rotate the coordinate system of the pinhole camera $\cam{P}$ such that $\V{q_{11}} = \begin{bmatrix} 1 & 0 & 1 \end{bmatrix}^\top$.

\vspace{4px}
\noindent \textbf{Null-space parameterization of $\M{H}_{\M{K}}$:}
\noindent Let us define the homography induced by the three 2D-3D point correspondences between the coordinate systems of $\cam{G}$ and $\cam{P}$ via point transfer through the plane $\pi$. Denoting the homography matrix as $\M{H} \in \mathbb{R}^{3 \times 3}$ we can write 
\begin{equation} \label{eq:H}
    \M{H} = \M{R} - \V{t} \V{N}^\top \enspace.
\end{equation} The constraints imposed by a 2D-3D point correspondence, of the form~\eqref{eq:3d2d}, become 
\begin{eqnarray}\label{eq:lin_const_on_H0}
    \M{H} \V X^{\g}_j = \alpha_j \M{K}^{-1} \V p_j \enspace, \enspace j=2,\dots4 \enspace.
\end{eqnarray} 
Let us write $\V{t}_{\M{K}} = \M{K} \V{t}$ and $\M{H}_{\M{K}} = \M{K} \M{H}$. Then Eqs.~\eqref{eq:H} and~\eqref{eq:lin_const_on_H0} can be rewritten as 
\begin{eqnarray}
\M{H}_{\M{K}} = \M{K} (\M{R} - \V{t} \V{N}^\top) =  \M{K} \M{R} - \V{t}_{\M{K}} \V{N}^\top \label{eq:HK_form}  \\
\M{H}_{\M{K}} \V X^{\g}_j = \alpha_j \V p_j \enspace, \enspace j=2,\dots4 \label{eq:lin_const_on_H} \enspace.
\end{eqnarray}
Eq.~\eqref{eq:lin_const_on_H} gives us six linear equations in the nine elements of $\M{H}_{\M{K}}$. From these equations we can obtain a three-dimensional nullspace parameterization  
of the matrix $\M{H}_{\M{K}}$. Thus, we can express  $\M{H}_{\M{K}}$ as a function of three unknown variables, $n_1, n_2$ and $n_3$.
In fact, we can express $\M{H}_{\M{K}}$ as 
\begin{eqnarray}\label{eq:3var_null_space_H}
 \M{H}_{\M{K}} = n_1 \M{N}_1 +n_2 \M{N}_2 +n_3 \M{N}_3 \enspace,
\end{eqnarray} 
where $ \M{N}_1, \M{N}_2 $ and $\M{N}_3$ are the matrix forms of the basis vectors of the null space.

\vspace{4px}
\noindent \textbf{Constraint on $\M{H}_{\M{K}}$:}
\noindent Now we derive constraints on the matrix $\M{H}_{\M{K}}$. 
We can rewrite Eq.~\eqref{eq:HK_form} as 
\begin{equation}\label{eq:R_as_fun_of_KHT}
    \M{R} = \M{K}^{-1}( \M{H}_{\M{K}} + \V{t}_{\M{K}} \V{N}^\top) \enspace.
\end{equation}
Since $\M{R} \in \bf{SO}(3)$, 
we have $\M{R} \M{R}^\top = \M{R}^\top \M{R} = \M{I}$. Thus we have the following set of constraints from Eq.~\eqref{eq:R_as_fun_of_KHT}: 
\begin{eqnarray}\label{eq:R_constraints_on_H}
  \M{K}^{-1} \left( \M{H}_{\M{K}} + \V{t}_{\M{K}} \V{N}^\top \right) \left( \M{H}_{\M{K}} + \V{t}_{\M{K}} \V{N}^\top  \right)^\top \M{K}^{-\top} = \M{I} \enspace, \nonumber \\
  ( \M{H}_{\M{K}} + \V{t}_{\M{K}} \V{N}^\top)^\top \M{K}^{-\top} \M{K}^{-1} ( \M{H}_{\M{K}} + \V{t}_{\M{K}} \V{N}^\top) = \M{I} \enspace.
\end{eqnarray} 
Due to the proposed coordinate system transform, the constraint~\eqref{eq:2d2d_elim_depths} for the 2D-2D point correspondence becomes
\begin{eqnarray} 
\label{eq:2d2d_H}
 \V{p}_1^\top  \left[\M{K} \M R  \V q_{11} \right]_{\times} \M{K} \V t =  0.
\end{eqnarray}
Substituting 
$\M R$ from Eq.~\eqref{eq:R_as_fun_of_KHT} into 
Eq.~\eqref{eq:2d2d_H}, we have 
\begin{eqnarray} \label{eq:2d2d_constraint_on_H}
  ( (\M{H}_{\M{K}} + \V{t}_{\M{K}} \V{N}^\top ) \V q_{11} )^\top \left[ \V{t}_{\M{K}} \right]_{\times} \V p_1  = 0 \enspace.
\end{eqnarray}
The equations in~\eqref{eq:R_constraints_on_H} and~\eqref{eq:2d2d_constraint_on_H} 
are the constraints on the unknown quantities, $\M{H}_{\M{K}}$, $\V{t}_{\M{K}}$, and $\M{K}$, and the known quantities, $\V{N} = \begin{bmatrix} 0&0&d \end{bmatrix}^\top$ and $\V{p_1} = \begin{bmatrix} x_1 & y_1 & 1 \end{bmatrix}^\top$.
Let us define the ideal generated by these equations as $I \subset \mathbb{C}\left[ \varepsilon \right] $~\cite{cox2006using}, where $\varepsilon$ contains the nine unknowns from $ \M{H}_{\M{K}}$, three from $\V{t}_{\M{K}}$, the inverse of the focal length $w= {1 \over f}$,  and $d$, $x_1$, and $y_1$. Note that $x_1$, $y_1$, and $d$ are known and here we treat them as known symbolic variables.
Now, we can use the elimination ideal technique~\cite{kukelova2017clever} to eliminate three unknowns of $\V{t}_{\M{K}}$ and $w$ from this ideal. 
\Ie, we compute an elimination ideal $I_1$ that contains polynomials only in nine unknown variables from $\M{H}_{\M{K}}$ and three known variables, $d$, $x_1$, and $y_1$. 
Note, that this elimination ideal can be computed offline using some algebraic geometry software like Macaulay 2~\cite{M2}. We found that such an elimination ideal is generated by eight polynomials (two of degree $3$, one of degree $4$, and five of degree $8$) in $12$ variables (nine unknown and three known). For more details on elimination ideals see~\cite{cox2006using,kukelova2017clever}. 

Substituting the three-variable parameterization of $\M{H}_{\M{K}}$ from Eq.~\eqref{eq:3var_null_space_H} into the generators of the ideal $I_1$, we obtain a system of eight equations in three unknowns. Using Macaulay 2~\cite{M2}, we verified that this system has up to $12$ solutions. This system defines the minimal formulation for the $\mathbf{H}13f$ case. We used two state-of-the-art algebraic methods, \ie, the \gb-based automatic generator\cite{larsson2017efficient} including the basis sampling strategy~\cite{larsson2018beyond} as well as the resultant-based generator~\cite{BhayaniKH2021}, to generate solvers for this system of eight equations in three unknowns. 
We found that smaller solvers can be obtained if, instead of using all eight equations, we use only six equations (two of degree $3$, one of degree $4$, and three of degree $8$).  
Using the 
\gb method with the basis sampling strategy~\cite{larsson2018beyond}, the generated solver was of size $77 \times 89$, while the one generated using the resultant-based method \cite{BhayaniKH2021} was of size $91 \times 91$.
We report these solver sizes in Tab.~\ref{tbl:solvers}. 

\vspace{4px}
\noindent \textbf{Extracting pose from $\M H$:}
\noindent Once we have computed the solutions to $\M{H}_{\M{K}}$, we can  estimate the value of 
$w=1/f$ from the Eq. \eqref{eq:R_constraints_on_H} through variable elimination and substitution. From $w$, we can then compute the calibration matrix $\M{K}$ and subsequently the homography matrix $\M{H} = \M{K}^{-1} \M{H}_{\M{K}}$.
An important step here is to efficiently extract the relative pose, \ie, the rotation matrix $\M{R}$ and the translation vector $\V{t}$ from $\M{H}$. Our chosen coordinate system transformation
plays a crucial role here. 
Substituting $\V{N} = \begin{bmatrix} 0 &0 &d \end{bmatrix}^\top$ in Eq. \eqref{eq:H}, we have 
\begin{equation}\label{eq:RT_from_H}
\M{R} = \M{H} + \V{t} \begin{bmatrix} 0 &0 &d \end{bmatrix} \enspace.
\end{equation}
Writing $\M{H} = \begin{bmatrix} \V{h}_1 & \V{h}_2 & \V{h}_3 \end{bmatrix}$, the rotation matrix $\M{R}$ can be computed as 
\begin{equation}
    \M{R} = \begin{bmatrix} \V{h}_1 & \V{h}_2 & \left[\V{h}_1 \right]_{\times} \V{h}_2 \end{bmatrix} \enspace.
\end{equation} 

\noindent Again, from Eq. \eqref{eq:RT_from_H} and the computed values of $\M{H}$,
the translation vector $\V{t}$ can be computed as 
\begin{equation}
    \V{t} = \dfrac{1}{d} (\left[ \V{h}_1 \right]_{\times} \V{h}_2 - \V{h}_3) \enspace.
\end{equation}


\subsection{$\mathbf{H}32f$}\label{subsec:h32f}
\noindent In the second configuration, we have three 2D-2D point correspondences $ \V p_j  \leftrightarrow (\V q_{ij}, \V t_{\g_i})$ and two 2D-3D point correspondences $ \V p_j \leftrightarrow \V X^{\g}_{j}$. Thus, there are three possible generalized camera configurations, \ie, $[1]$, $[2]$ and $[3]$.
For all three camera configurations, we obtained the smallest solvers using the \textbf{\textit{Rotation $\&$ translation}} parameterization. Such a parameterization for $\mathbf{H}32f$ was already proposed in 
\cite{JosephsonBKA2007}, where the authors suggested to use a coordinate transformation resulting in a system of five equations in five variables with $52$ solutions.
This system is, however, complex and it leads to a huge solver\footnote{This solver was not presented in~\cite{JosephsonBKA2007} and only the parameterization was discussed.}. In order to generate a feasible solver with fewer solutions, in this paper, we reparameterized the product $\M K \M R$. Next, we review the coordinate transformation and the parameterization proposed in \cite{JosephsonBKA2007} and then describe our proposed solver. We obtain the same solver for all three configurations $[1],[2]$ and $[3]$.

\vspace{4px}
\noindent \textbf{Coordinate system transform:}
\noindent W.l.o.g., we can translate and rotate the coordinate system of $\cam{G}$ 
such that $\V{X}_{4}^{\g} = \begin{bmatrix} 0 & 0 & 1  \end{bmatrix}^\top$ and $\V{X}_{5}^{\g} = \begin{bmatrix} 0 &  0  & 0 \end{bmatrix}^\top$. Moreover, we can rotate the coordinate system of the pinhole camera $\cam{P}$ such that $\V{p_{5}} = \begin{bmatrix} 1 & 0 & 1 \end{bmatrix}^\top$.

By this coordinate system transform, the constraint~\eqref{eq:3d2d} for the fifth point correspondence
$ \V p_5 \leftrightarrow \V X^{\g}_{5}$ leads to 
\small \begin{equation}
 \M R \begin{bmatrix} 0 \\ 0 \\ 0 \end{bmatrix} +\V t =  \alpha_{5} \M{K}^{-1}  \begin{bmatrix} 1 \\ 0 \\ 1 \end{bmatrix} \implies \M{K} \V{t} = \alpha_{5}  \begin{bmatrix} 1 \\ 0 \\ 1 \end{bmatrix} \enspace.
\end{equation} \normalsize
Substituting the above expression of $\M{K} \V{t}$ in the constraints for the 2D-2D point correspondences and the fourth 2D-3D point correspondence, we obtain 
\small \begin{eqnarray}
 (\M{K} \M{R} \V{p}_j)^\top \left[  \V{q}_{ij}  \right]_{\times} (\M{K} \M{R} \V{t}_{g_i} +  \alpha_5 \begin{bmatrix} 1 \\ 0 \\ 1 \end{bmatrix}) &=& \V{0} \enspace, \label{eq:hyb32_2d2d_telim_reparm} \\
  \left[ \V{q}_4\right]_{\times} ( \M{K} \M{R} \V{X}_4 +  \alpha_5 \begin{bmatrix} 1 \\ 0 \\ 1 \end{bmatrix} ) &=& \V{0} \enspace,  \label{eq:hyb32_2d3d_telim_reparm}
\end{eqnarray}\normalsize  where $i \! \in\! \lbrace 1,2,3\rbrace$ and $j \! =\!  1,\dots,3$. The constraints~\eqref{eq:hyb32_2d2d_telim_reparm} and~\eqref{eq:hyb32_2d3d_telim_reparm} together give us a set of five equations in the quantities $\M K$, $\M R$, and $\alpha_5$. We parameterized the rotation matrix $\M R$ using quaternions. We have five unknowns,
one for the calibration matrix $\M K$, three for the rotation matrix $\M R$ and one for $\alpha_5$. Using the automatic generator \cite{larsson2017efficient,larsson2018beyond}, we obtained a large solver of size $1866 \times 1918$ and with $52$ solutions.

\vspace{4px}
\noindent \textbf{Our approach:}
\noindent 
To simplify this solver, we reparameterize the  product $\M K \M R $ as a function of four new variables, $r_1, r_2, r_3$ and $r_4$. This reparameterization is described in the SM. 
With this reparameterization, the constraints~\eqref{eq:hyb32_2d2d_telim_reparm} and~\eqref{eq:hyb32_2d3d_telim_reparm} give us a set of five equations in five variables $r_1, r_2, r_3, r_4$, and $\alpha_5$. The two linearly independent equations from~\eqref{eq:hyb32_2d3d_telim_reparm} are linear in $r_1$ and $r_2$. Hence both these variables can be expressed as functions of the other three variables, $r_3, r_4$, and $\alpha_5$. Substituting these expressions of $r_1$ and $r_2$ in Eq.~\eqref{eq:hyb32_2d2d_telim_reparm}, we obtain a system of three equations in three variables $r_3, r_4$, and $\alpha_5$, each of degree eight. Let us denote the polynomial set as $E = \lbrace e_1, e_2, e_3 \rbrace$. The ideal $I \in  \mathbb{C}\left[ r_3, r_4, \alpha_5 \right]$ generated by $E$ is not zero-dimensional~\cite{cox2006using}. Specifically, if $\alpha_5 = 0 \ \&\ r_3^2+r_4^2-1=0$, or $r_3^2+r_4^2=0$, we have a set of trivial solutions to $E=0$\footnote{The resulting possible degeneracies for $\alpha_5 = 0 \ \&\ r_3^2+r_4^2-1=0$ and $r_3^2+r_4^2=0$ can be avoided by a random rotation of the camera $\cam{P}$.}.

In order to generate a solver from $E$, we have to remove those solutions of $E=0$ where $\alpha_5=0 \ \& \  r_3^2+r_4^2-1=0$ or $r_3^2+r_4^2=0$. This can be achieved by saturating  the ideal $I$ \wrt $\alpha_5$ and $r_3^2+r_4^2$~\cite{cox2006using}. This saturation results to a solver of size $475 \times 501$ generated using the 
generator~\cite{larsson2017efficient,LarssonAO2017}.

To generate a smaller solver, we want to avoid saturating \wrt $\alpha_5$ and also avoid adding an extra variable. Thus, we augment $E$ with extra polynomials which vanish on all non-trivial solutions to $E=0$ but not if $\alpha_5=0 \ \& \ r_3^2 + r_4^2-1=0$. We next show how to generate such polynomials. The form of the $i$-th polynomial in $E$ is 
 \begin{eqnarray}
 e_i = ( r_3^2+r_4^2-1) \phi_i +\alpha_5 \psi_i \enspace,
 \end{eqnarray} where $\phi_i$ and $\psi_i$ are polynomials in $r_3, r_4, \alpha_5$. Then, the set of equations $E=0$ can be written in matrix form as 
 \begin{eqnarray}
 \M{M} \V{b} = \begin{bmatrix} \phi_1 & \psi_1 \\ \phi_2 &  \psi_2 \\ \phi_3 & \psi_3 \end{bmatrix}_{3 \times 2} \begin{bmatrix} r_3^2+r_4^2-1 \\ \alpha_5 \end{bmatrix}_{2\times 1} = \V{0}_{3\times 1} \enspace.
 \end{eqnarray}
 The determinant of each $2 \times 2$ submatrix of $\M M$ vanishes only for those solutions of $E=0$ such that $\alpha_5\neq 0$ or $r_3^2+r_4^2-1\neq 0$. We have a total of three such determinant expressions, each a polynomial of degree ten in the variables $r_3, r_4$, and $\alpha_5$, out of which two are linearly independent. 
 Let us denote the augmented system consisting of $E$ and two of these polynomials as $E_a$.  
 This system has $26$ solutions and does not contain solutions where $\alpha_5 = 0\ \& \ r_3^2+r_4^2-1= 0$. 
 The ideal $I_a$ generated by $E_a$ now has trivial solutions only if $r_3^2+r_4^2=0$. Therefore, we saturate $I_a$ w.r.t. $r_3^2+r_4^2$ and use \gb-based method with heuristic sampling~\cite{larsson2018beyond} to obtain a solver of size $289 \times 315$. We failed to generate a solver using the resultant-based approach~\cite{BhayaniKH2021}. 

\subsection{$\mathbf{H}51f$}
\noindent In this scenario, we have five 2D-2D point correspondences $ \V p_j \leftrightarrow (\V q_{ij}, \V t_{\g_i})$ and one 2D-3D point correspondence $\V p_6 \leftrightarrow  \V X^{\g}_{6} $, which together give us a set $F$ of $7$ equations. For this case we have $5$ possible generalized camera configurations, \ie, $ [1]$, $\dots$, $[5]$. We first study the first four configurations. 

To solve these configurations we again  tested three different groups of parameterizations, \ie, 
$\M R \&  \V t$, $\M H$, and $\M F$
together with different simplification, reparameterizations, eliminations and solution strategies.
It turns out that the configurations $[1]$, $[2]$, $[3]$, and $[4]$ are the most challenging configurations among all those studied in this work. For these problems the smallest solvers were generated using the  \textbf{\textit{Rotation $\&$ translation}} parameterization. However, even  after different reparameterizations and simplifications these solvers 
 are huge.
 The sizes of the elimination template matrices for the smallest obtained solvers for these four camera configurations of $\mathbf{H}51f$ are listed in Tab.~\ref{tbl:solvers}. More details about these solvers and parameterizations can be found in the SM.

\vspace{-1.6ex}
\paragraph{$\mathbf{H}51f[5]$:}
In contrast to configurations $[1],\dots,[4]$, if all $5$ correspondences are detected by the same camera pair, we have 
the simplest scenario.
In this case, the problem is reduced to the problem of estimating the relative pose between two pinhole cameras from $6$ 2D-2D point correspondences, where the sixth correspondence is obtained by projecting the 3D point into the camera $\cam{G}_i$. This is followed by the scale estimation from the remaining constraint given by the 2D-3D point correspondence.
Such a solution was already proposed in~\cite{Josephson09CVPR}. 
However, there the authors suggested to use the 6pt solver~\cite{Stewenius2005} for two-sided common focal length and therefore they reported $15$ solutions. This configuration is not practical since $\cam{G}$ is usually calibrated and moreover, it usually doesn't have the same focal length as the query 
camera. In \cite{Josephson09CVPR}, the authors did not test the proposed solver.

In this paper, we consider a more practical scenario where $\cam{G}$ is calibrated. This results in estimating the essential matrix and the focal length of $\cam{P}$ from six 2D-2D correspondences, known as the one-sided focal length problem~\cite{bujnak20093d}. This problem has nine solutions.
To solve this problem we use the \gb-based solver proposed in~\cite{kukelova2017clever}, which eliminates the unknown focal length using the elimination ideal method and thus results in a smaller elimination template matrix of size $9 \times 18$ compared to~\cite{bujnak20093d}. Using the resultant-based generator~\cite{BhayaniKH2021}, we obtained a solver of size $10 \times 10$. 

\newcommand{\fst}[1]{{\bf #1}}
\newcommand{\snd}[1]{{\underline{#1}}}
\newcommand{\dname}[1]{\scalebox{0.8}{#1}}
\begin{table}[t]
    \centering
    \resizebox{\columnwidth}{!}{
    \begin{tabular}{l c rl c rl c rl c rl c c} \toprule
 && \multicolumn{2}{c}{\dname{KingsCollege}} && \multicolumn{2}{c}{\dname{OldHospital}} && \multicolumn{2}{c}{\dname{ShopFacade}} &&  \multicolumn{2}{c}{\dname{StMarysChurch}} &&  \\ \cmidrule{3-4} \cmidrule{6-7} \cmidrule{9-10} \cmidrule{12-13}

Solver &&  & +LO &&  & +LO &&  & +LO &&  & +LO && Runtime \\ \midrule
$\mathbf{P4P}f$                 &&      41.1 & \snd{70.3}&&      33.0 &      52.2 &&      79.6 &      89.3 &&      59.2 & \fst{83.8} &&  \fst{0.06} \\
$\mathbf{P3.5P}f$                &&      50.4 &      69.4 &&      36.8 & \snd{55.5}&&      83.5 & \fst{95.1}&&      72.1 &      82.6 &&  \snd{0.08} \\
$\mathbf{H}13f$                && \fst{61.5}&      69.4 && \snd{45.1}& \fst{57.7}&& \fst{92.2}& \fst{95.1}&& \snd{76.4}& \fst{83.8} &&   0.75 \\
$\mathbf{H}51f[5]$                 &&      42.9 & \fst{70.8}&&      36.3 &      46.7 &&      82.5 &      89.3 &&      62.3 & \snd{82.8} &&   0.19 \\
$\mathbf{P35P}f+\mathbf{H}13f+\mathbf{H}51f[5]$      && \snd{60.9}&      69.4 && \fst{47.3}& \snd{55.5}&& \snd{86.4}& \snd{90.3}&& \fst{77.4}&      83.2 &&   0.75 \\
\bottomrule
    \end{tabular}}
    \vspace{2pt}
     \caption{\textbf{Hybrid Localization on Cambridge Landmarks.} The table shows the percentage of camera poses localized within 1$^\circ$ and 0.5m. In the table we show the results of both vanilla RANSAC and LO-RANSAC. For a fair comparison the model scoring in RANSAC is the same for all methods (taking into account both 2D-2D and 2D-3D correspondences). The best result is highlighted in bold, and the second-best is underlined. Table also shows the median runtime for RANSAC in seconds.}
    \label{tbl:cambridge_landmarks}
\end{table}
\begin{figure*}[t]
\centering
    \subfloat[]{\includegraphics[width=0.29\linewidth]{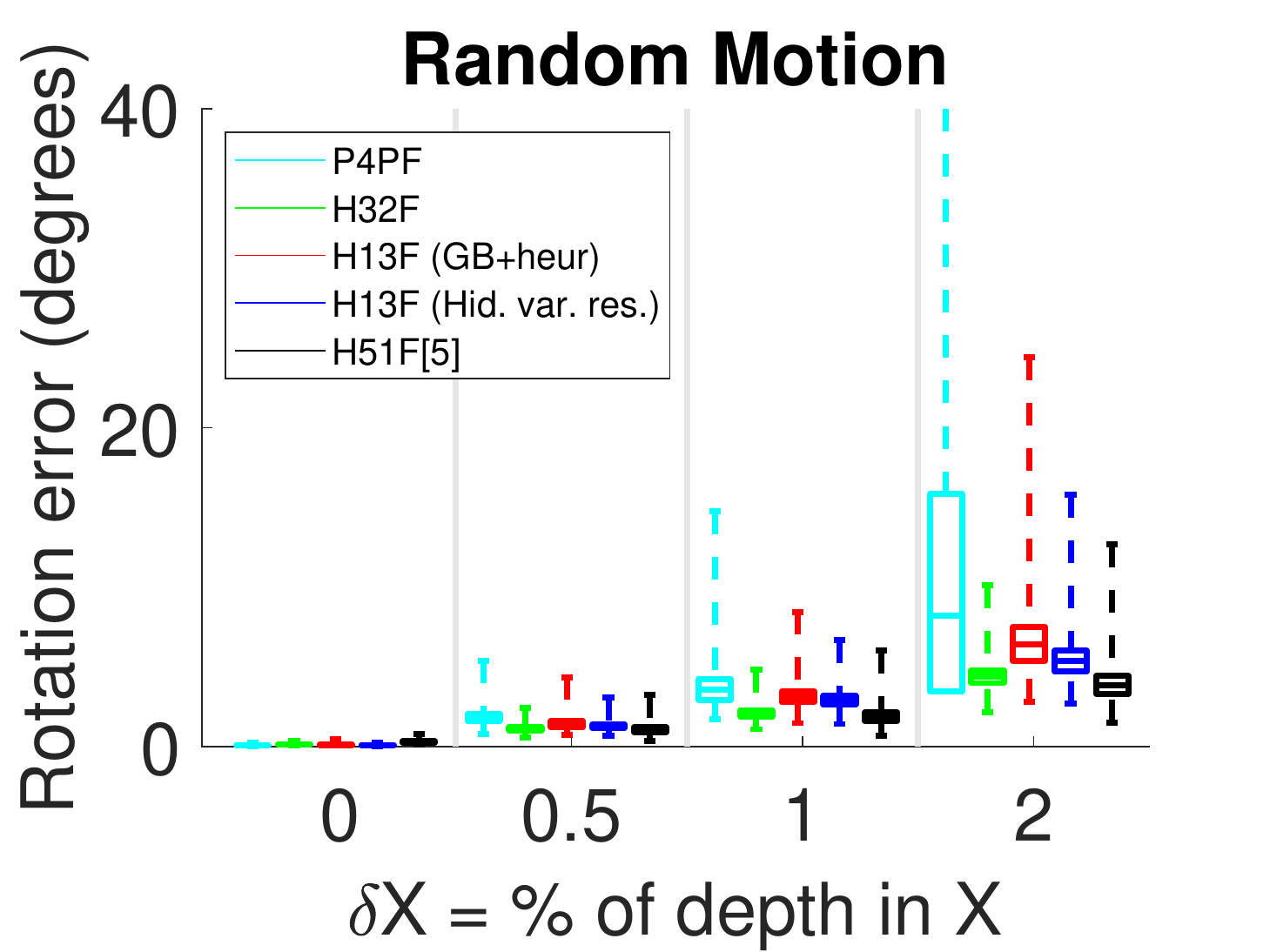}}
    \subfloat[]{\includegraphics[width=0.29\linewidth]{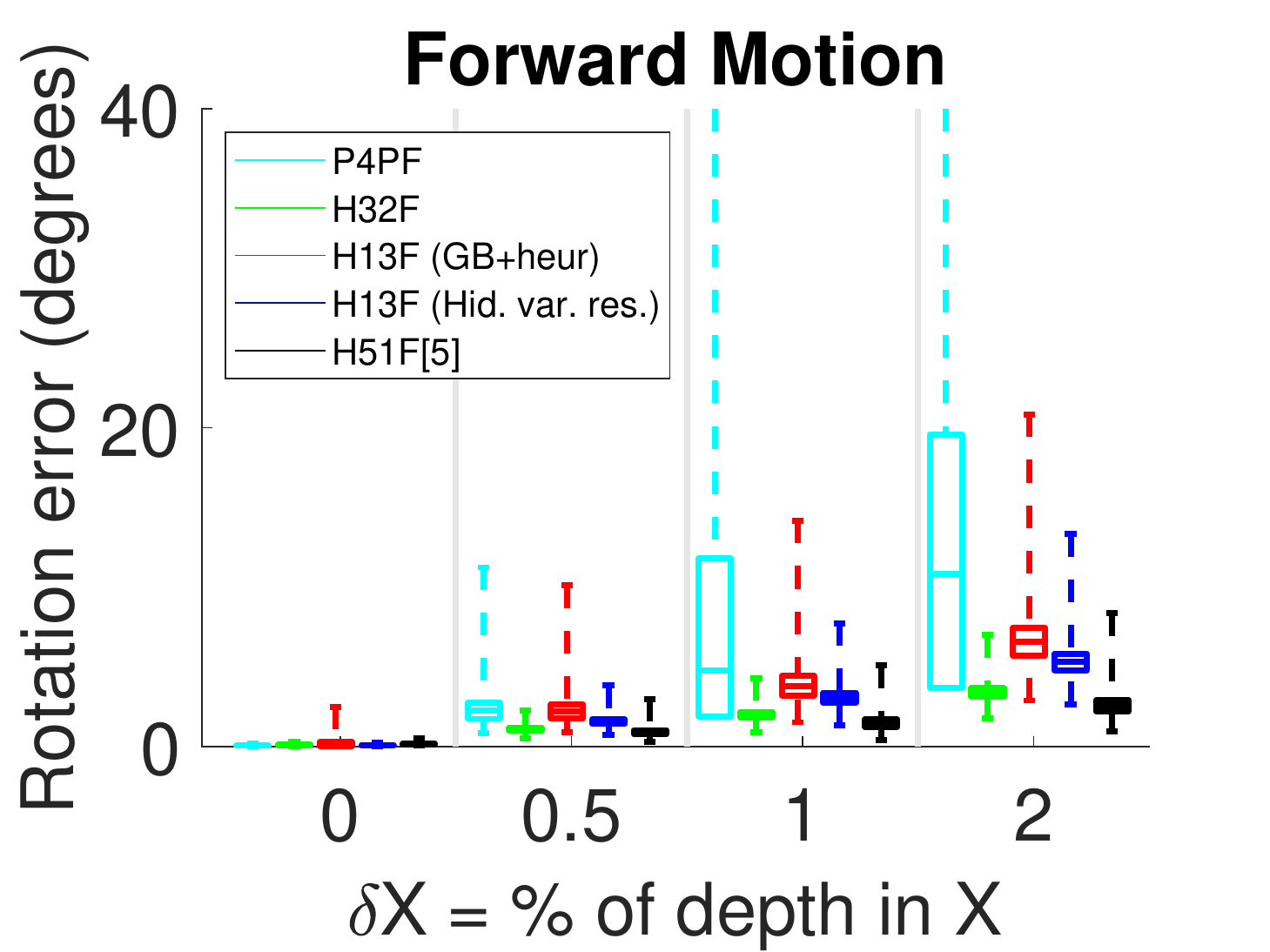}}
     \subfloat[]{\includegraphics[width=0.29\linewidth]{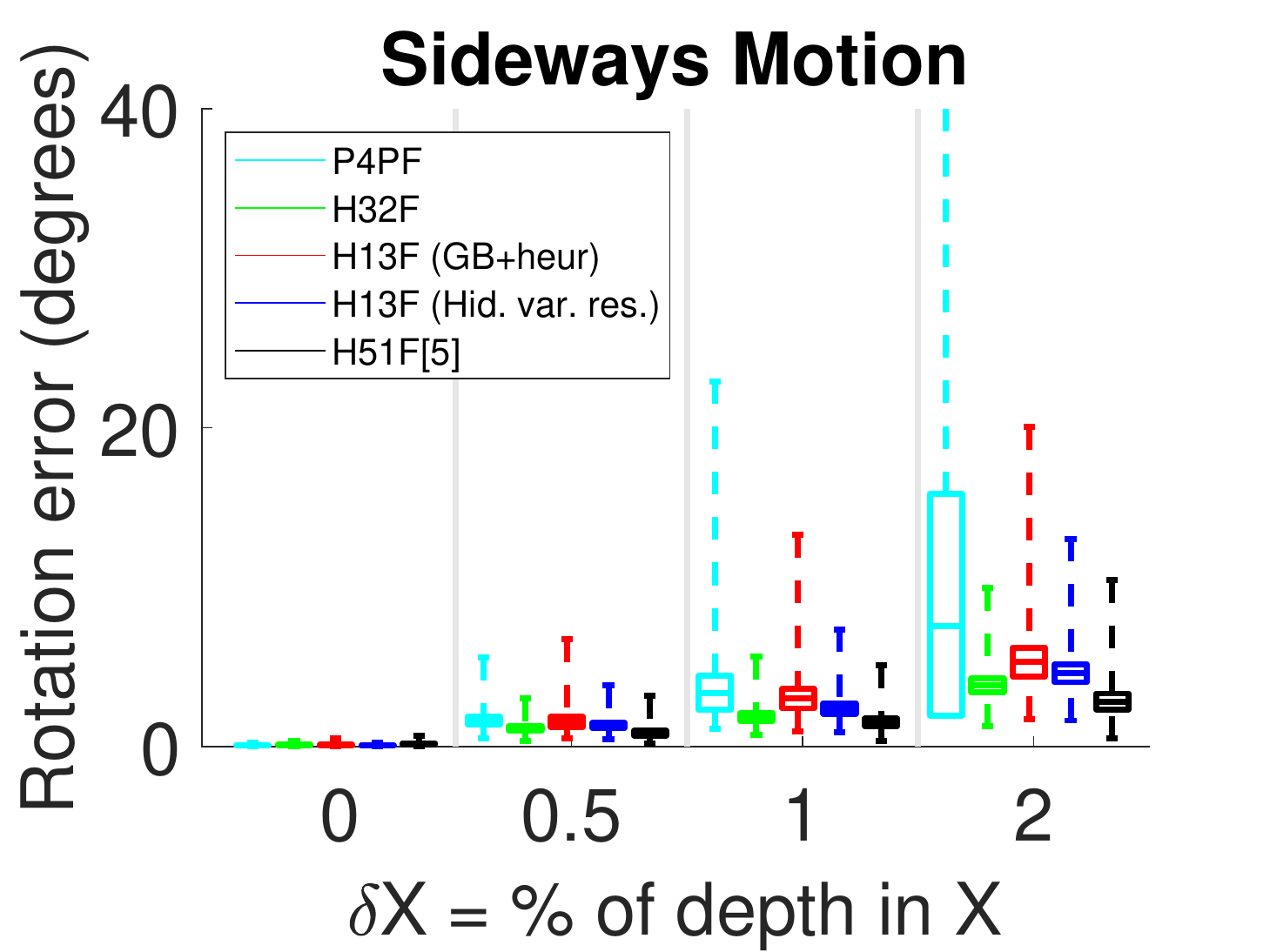}}
     
    \subfloat[]{\includegraphics[width=0.29\linewidth]{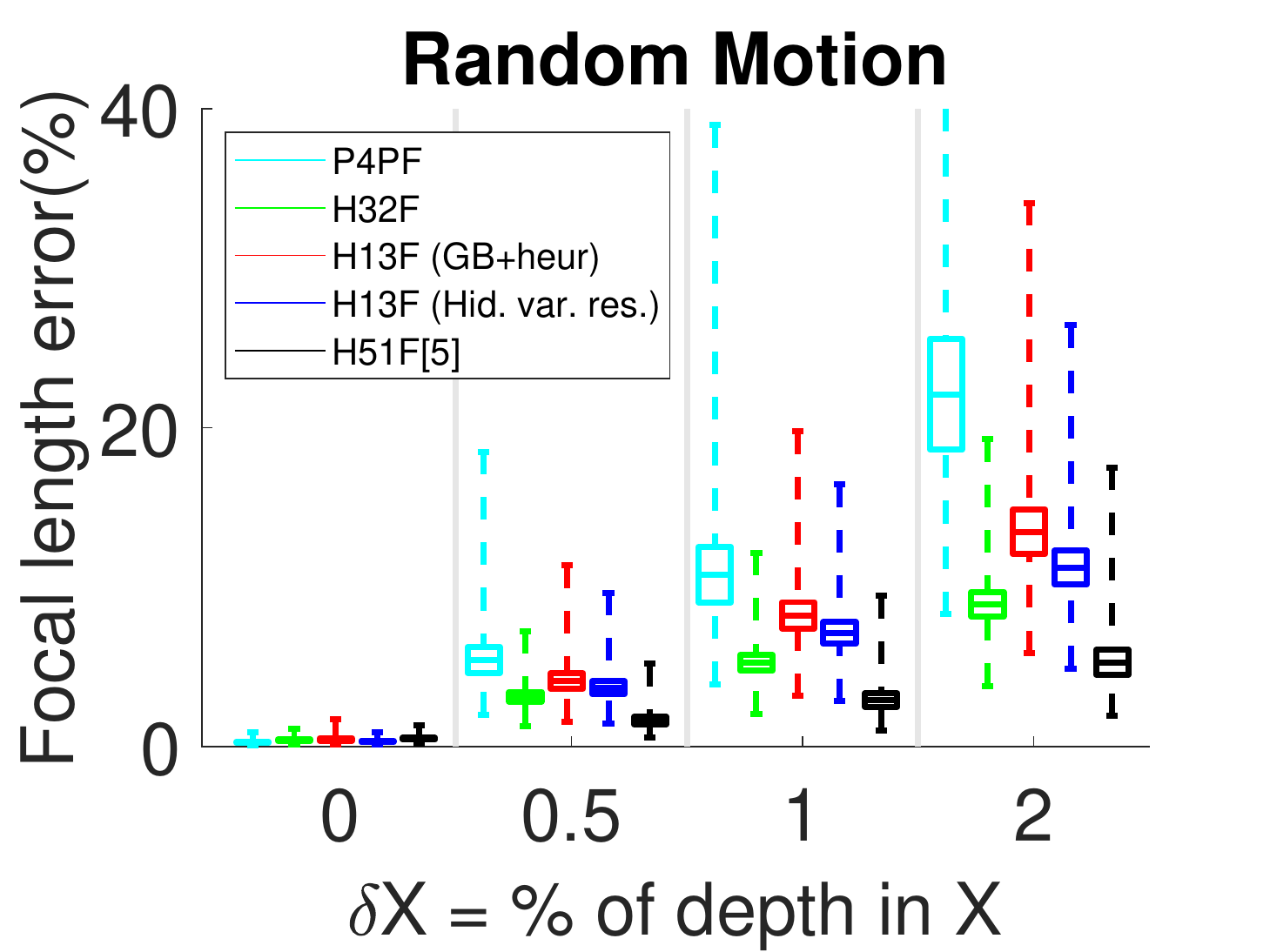}}
      \subfloat[]{\includegraphics[width=0.29\linewidth]{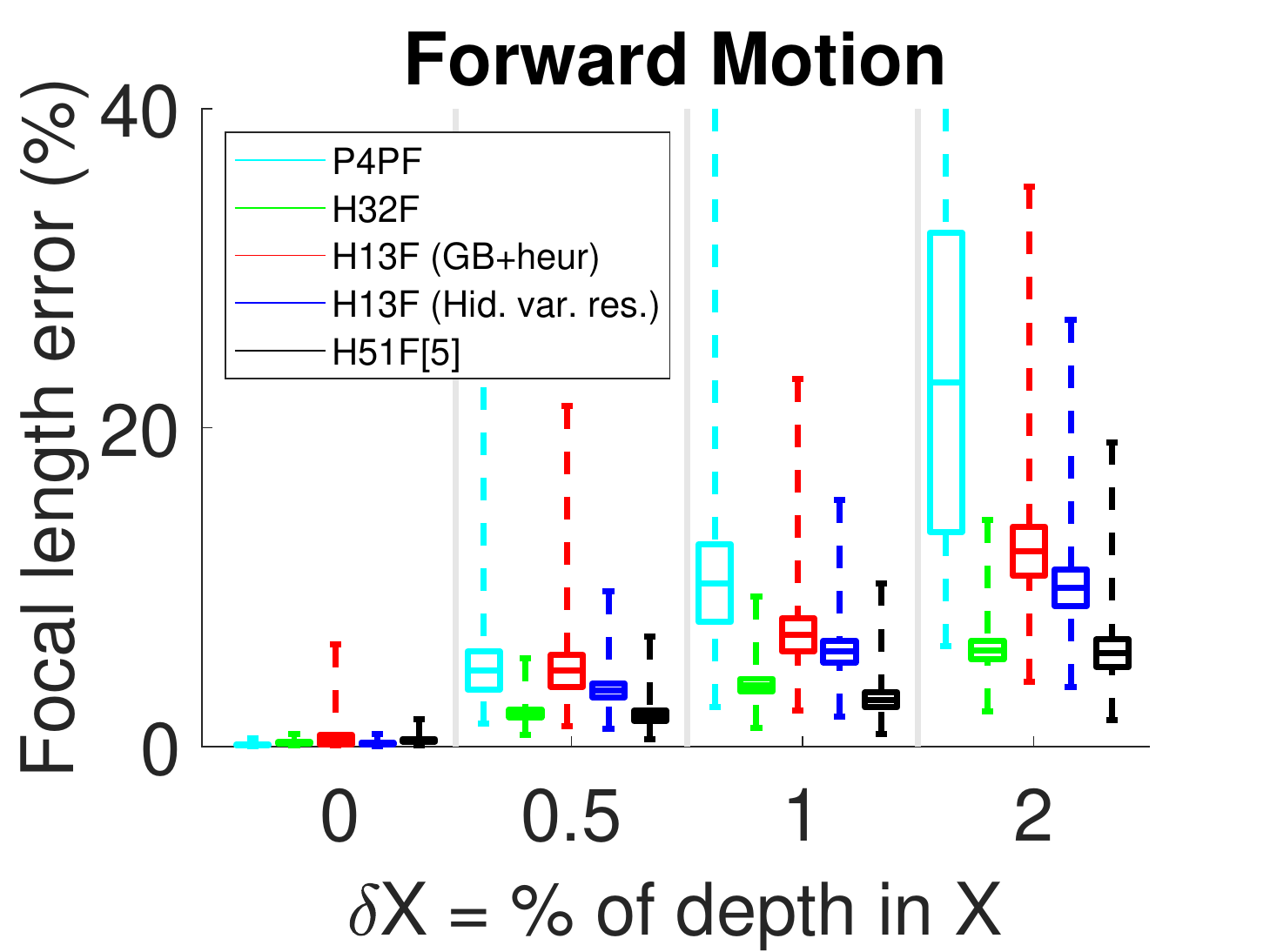}}
    \subfloat[]{\includegraphics[width=0.29\linewidth]{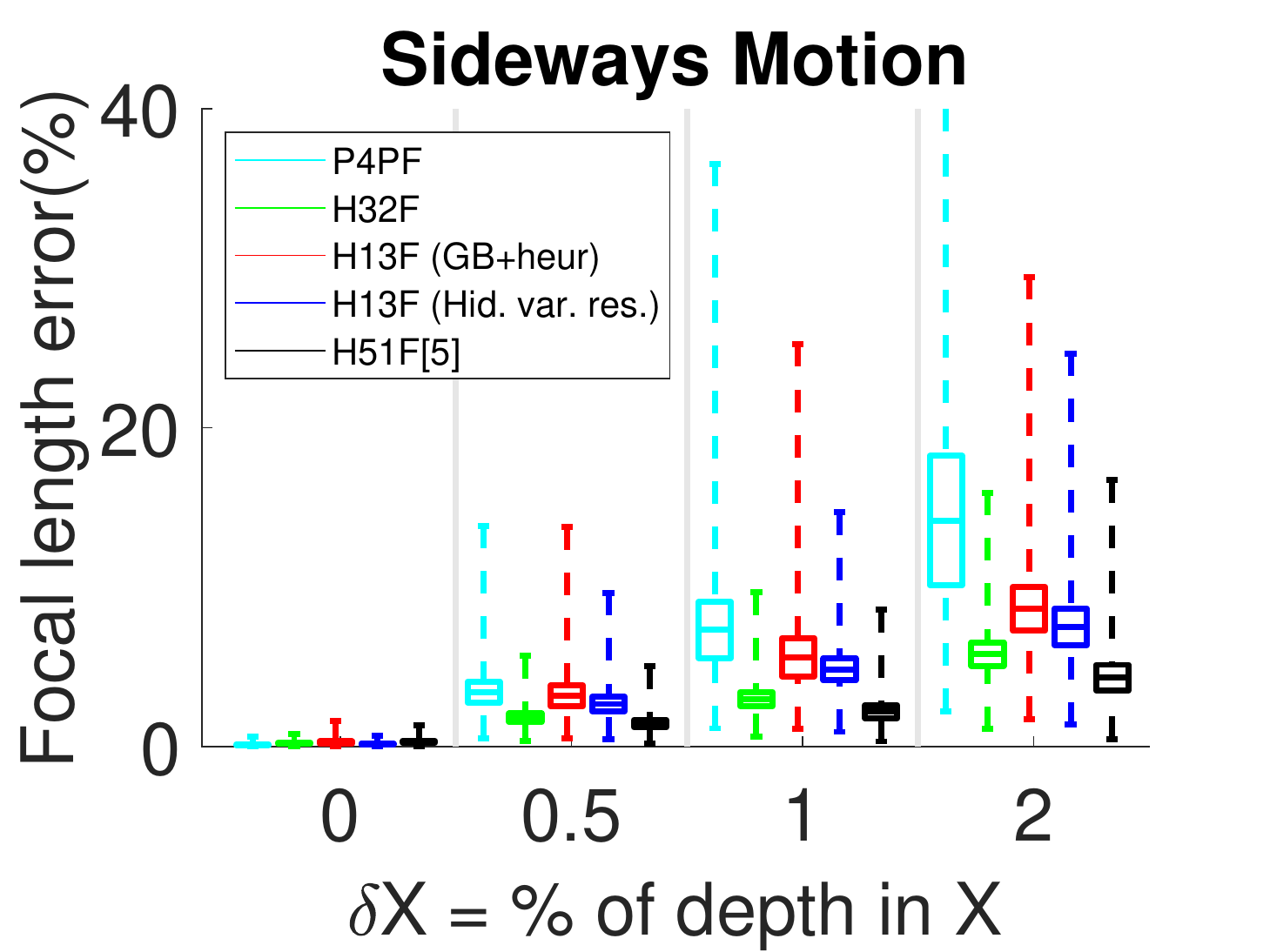} }
     
      \subfloat[]{\includegraphics[width=0.29\linewidth]{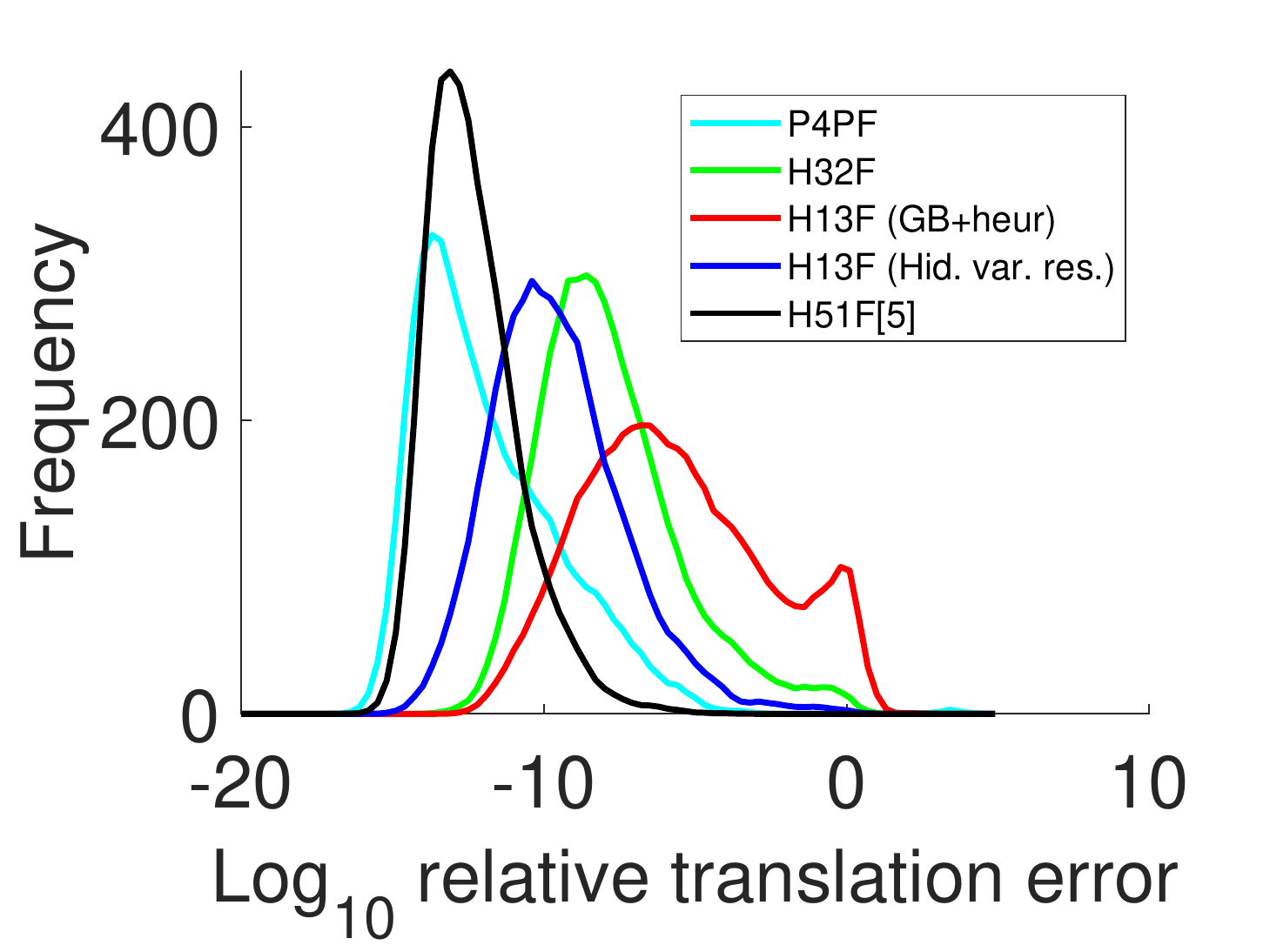} }
      \subfloat[]{\includegraphics[width=0.29\linewidth]{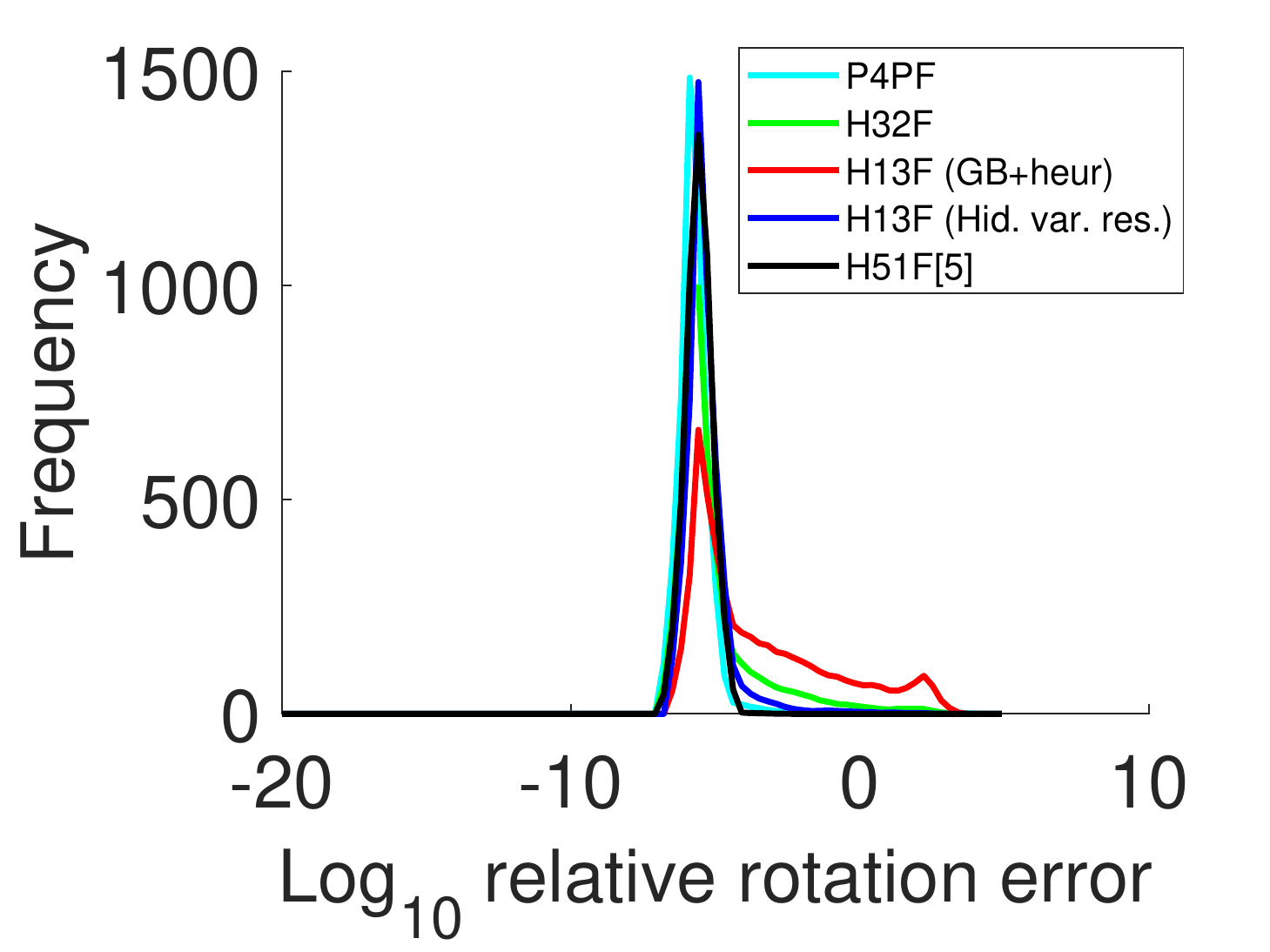}}
    \subfloat[]{\includegraphics[width=0.29\linewidth]{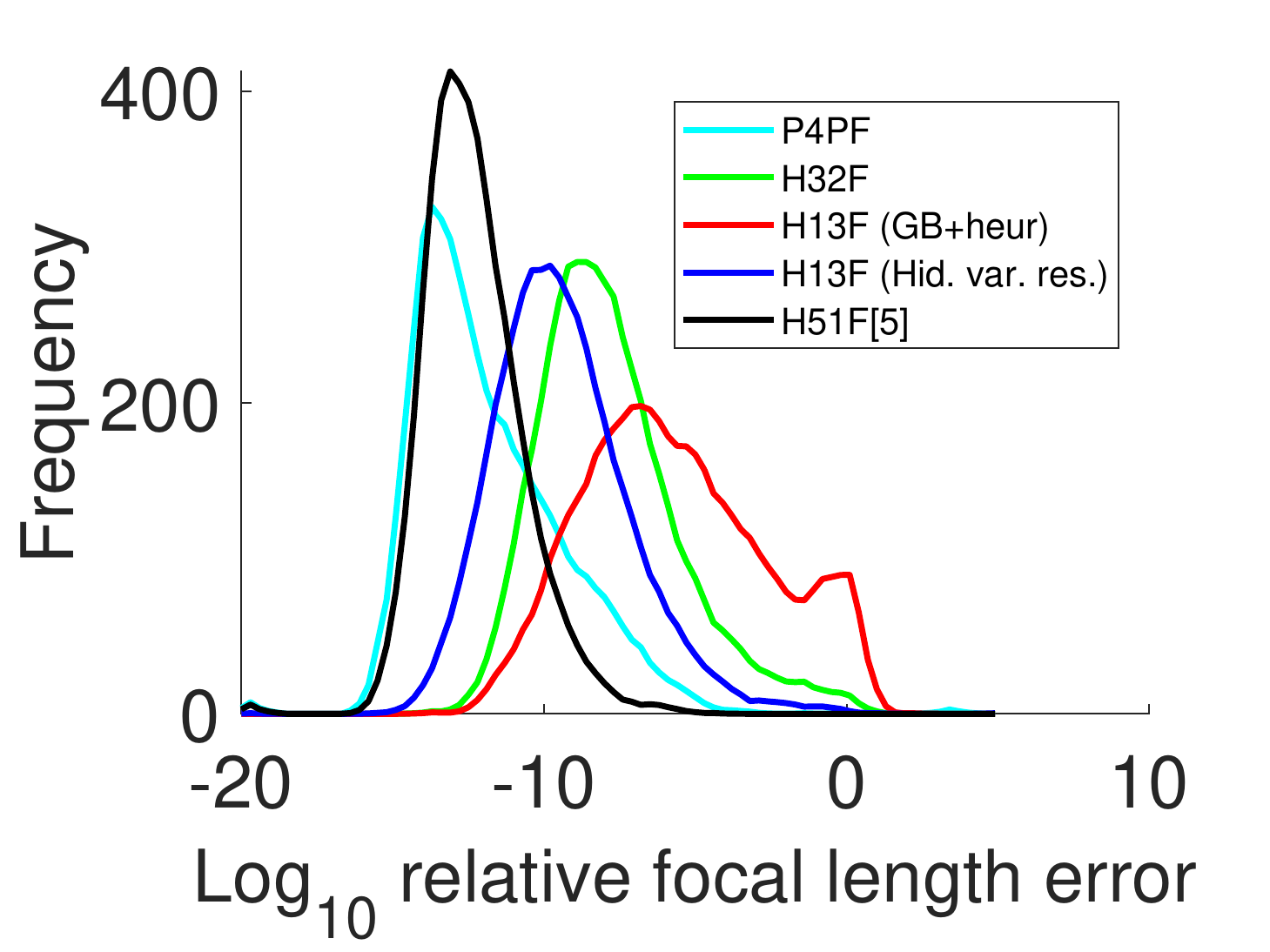} }
  
     \caption{ Error in rotation (\textbf{Row} 1) and focal length (\textbf{Row} 2)
     in the presence of increasing 3D point noise. Three camera motions considered : random motion \textbf{(a,d)}, forward motion \textbf{(b,e)}, and sideways motion \textbf{(c,f)}. Numerical stability of solvers for noiseless data and random motion  (\textbf{Row} 3). Our hybrid solvers are more robust to 3D point noise than the $\mathbf{P4Pf}$ absolute pose solver.}

\label{fig:synthgraphs}
\end{figure*}

\section{Synthetic experiments} \label{sec:synth}
\noindent For our synthetic scene tests, we generated 5K 3D scenes with known ground truth parameters. In each scene, the 3D points were randomly distributed within a cube of size $10 \times 10 \times 10$. Each 3D point was projected on up to $6$ cameras with realistic focal lengths. Five of these cameras represented the generalized camera $\cam{G}$ and one camera was considered as the pinhole camera $\cam{P}$. 
The orientations and positions of the cameras were selected at random such that they looked towards the origin from a random distance, varying from $15$ to $25$, from the scene. 
Images had the resolution of $1000 \times 1000$~px. 
We added Gaussian noise to the positions of the 3D points with the standard deviation $\sigma$ varying as a $\%$ of their depths to simulate the different quality of keypoints used for the triangulation of these 3D points. To simulate the noise in 2D-2D correspondences, we added $2$ px image noise. 
More experiments, \eg, for increasing image noise and fixed 3D point noise are in the SM.

We evaluated the stability of the proposed  $\mathbf{H}32f$, $\mathbf{H}13f$ and the $\mathbf{H}51f[5]$ solvers \wrt the SOTA absolute pose solver $\mathbf{P4Pf}$ \cite{kukelova2016efficient}. The graphs for synthetic experiments with increasing noise in the 3D points are provided in Fig. \ref{fig:synthgraphs}. 
We considered $3$ different camera motions in our tests, \ie, random, forward and sideways.
For each motion, we measured the error in the estimated rotation $\M R$, translation $\V t$ (in SM), and the focal length $f$, by varying the amount of noise in the 3D points. Note that we tested two derivations of our proposed $\mathbf{H}13f$ solver, the one based on the \gb \cite{larsson2018beyond} and one based on resultants \cite{BhayaniKH2021}. Both solvers have different numerical properties and sizes. Note from Fig. \ref{fig:synthgraphs}, that in the presence of increasing 3D point noise, for all three motions, the \gb-based $\mathbf{H}51f[5]$ solver and the proposed $\mathbf{H}32f$ and $\mathbf{H}13f$ solvers (both \gb-based and resultant-based) have much better stability than the SOTA $\mathbf{P4Pf}$ absolute pose solver, with the $\mathbf{H}51f[5]$ solver slightly outperforming the $\mathbf{H}32f$  and $\mathbf{H}13f$ solvers. 
The reason for this behavior is that our hybrid solvers are not only using 2D-3D correspondences, unlike the 
$\mathbf{P4Pf}$ solver. We also observe that the resultant-based solver for $\mathbf{H13}f$ has similar or slightly better stability compared to that of the \gb-based solver for $\mathbf{H13}f$.

\section{Real experiments} \label{sec:real_exp}
\noindent We evaluate the proposed $\mathbf{H}13f$ and $\mathbf{H}51f[5]$ solvers in a hybrid localization framework and consider four scenes from the Cambridge~Landmarks~\cite{Kendall2015ICCV} dataset. We did not test our proposed $\mathbf{H}32f$ solver as its template size makes it less practical for real-time use as compared to the other proposed solvers. 
For each query image, we establish tentative 2D-2D correspondences to the top-20 retrieved map images based on the DenseVLAD~\cite{Torii-CVPR15} image descriptor. From transitive matching we take all triangulated points as 2D-3D correspondences, and additionally add all 2D-2D correspondences that are either not triangulated, or have a track length less than 5 (as these are potentially less certain 3D points). We apply the solvers in the hybrid LO-RANSAC from~\cite{camposeco2018hybrid}, minimizing reprojection error (2D-3D) and Sampson error (2D-2D). We compare with the point-based solvers $\mathbf{P4P}f$ \cite{kukelova2016efficient} and $\mathbf{P3.5P}f$~\cite{larsson2017efficient}, as well as employing all solvers together in the hybrid framework from~\cite{camposeco2018hybrid}. Tab.~\ref{tbl:cambridge_landmarks} shows the percentage of queries within 1$^\circ$ and 0.5m. To highlight the differences between the accuracy of the solvers, we show the results both with and without local refinement in RANSAC. It can be seen that the proposed $\mathbf{H}13f$ solver is least affected by noise and returns the most accurate solutions. 

\section{Conclusion}
\noindent In this paper, we studied the challenging problem of estimating the semi-genera-lized pose from hybrid point correspondences for partially-calibrated cameras. 
By testing different parameterizations, elimination techniques and  solution strategies, solvers to all minimal configurations of 2D-2D and 2D-3D correspondences, \ie, $\mathbf{H}13f$, $\mathbf{H}32f$ and $\mathbf{H}51f$, and all possible camera configurations within the generalized camera, are derived.
The most practical solvers are evaluated  on synthetic and real scenes, showing the benefits of hybrid estimation compared to classical 2D-3D 
approaches. 
Our solvers fill gaps in the arsenal of minimal solvers and can be used inside hybrid RANSAC~\cite{camposeco2018hybrid}.


\section{Acknowledgements} 
Torsten Sattler was supported by the EU Horizon 2020 project RICAIP (grant agreement No. 857306) and the European Regional Development Fund under project IMPACT (No. CZ.02.1.01/0.0/0.0/15$\_$003/0000468). Zuzana Kukelova was supported by the OP VVV funded project CZ.02.1.01/0.0/0.0/16$\_$019/0000765 “Research Center for Informatics”. Viktor Larsson was supported by the strategic research project ELLIIT.

\clearpage
\bibliographystyle{ieee}
\bibliography{torsten,egbib}
\end{document}